\documentclass{bmvc2k}

\title{\textit{LEGAU:} Learning Semantic Gaussian Priors for Scalable Category-level Pose Estimation
}

\addauthor{Hongli Xu}{hongli.xu@tum.de}{1}
\addauthor{Zhaowei Lu}{zhaowei.lu@tum.de}{1}
\addauthor{Junwen Huang}{junwen.huang@tum.de}{1,3}
\addauthor{Jiaqi Hu}{jiaqi.hu@tum.de}{1,2}
\addauthor{Peter KT Yu}{peterkty@gmail.com}{4}
\addauthor{Benjamin Busam}{b.busam@tum.de}{1,3}
\addauthor{Federico Tombari}{tombari@in.tum.de}{1}
\addauthor{Slobodan Ilic}{slobodan.ilic@tum.de}{1,2}

\addinstitution{
Technical University of Munich\\
Munich, Germany
}

\addinstitution{
Siemens AG\\
Munich, Germany
}

\addinstitution{
Munich Center for Machine Learning\\
Munich, Germany
}

\addinstitution{
ROBOX\\
Cambridge, MA, USA
}

\runninghead{XU ET AL.}{LEGAU: Scalable Category Pose Estimation}

\usepackage{hyperref}

\usepackage{orcidlink}
\usepackage{graphicx}
\usepackage{booktabs}
\definecolor{cvprblue}{rgb}{0.21,0.49,0.74}

\usepackage{multirow}
\usepackage{arydshln} 
\usepackage[table]{xcolor} 
\usepackage{makecell}
\usepackage{colortbl}
\usepackage{pifont}
\usepackage{amssymb}   % \checkmark
\usepackage{pifont}    % \ding{55} 用于叉号
\usepackage{comment}
\usepackage{float}
\usepackage{placeins}   % 导言区
\newcommand{\cmark}{\ding{51}} % ✓
\newcommand{\xmark}{\ding{55}} % ✗
\usepackage[accsupp]{axessibility}

\newcommand{\ours}{\textit{LEGAU}}
\definecolor{hongli}{RGB}{0,0,0} % black
\newcommand{\hongli}[1]{{\color{hongli}#1}}

\begin{document}

\maketitle

\begin{abstract}
% Category-level 6D pose estimation remains challenging due to the difficulty of aligning object geometry and semantics across instances, which limits generalization to unseen shapes and domains. We present \textbf{\ours} as a unified framework that incorporates the Semantic Gaussian Field to enable scalable and generalizable pose estimation.
% Conditioned on a categorical text embedding, \ours
% processes an RGB-D image through a transformer-based fusion module that integrates visual, geometric, and semantic cues, decoding both the NOCS map and the complete 3D shape of the object. \ours reconstructs object geometry as a Semantic Gaussian Field that encodes both visual appearance and its internal semantics. This latent field serves as an implicit prior, guiding the pose alignment toward geometric and semantic consistency on unseen objects.
% Extensive experiments on synthetic and real-world benchmarks demonstrate strong generalization across diverse categories. \hongli{In particular, \ours{} improves pose estimation performance by up to \textbf{+22\%} on SOPE dataset, comparing to the state-of-the-art results.} These results highlight the effectiveness of jointly learning geometry and semantics within a unified field representation for robust category-level pose estimation.
Category-level 6D pose estimation from a single RGB-D observation is inherently under-constrained, since partial visible geometry must be interpreted together with a canonical object structure before a stable pose can be determined. We present \ours{}, a unified framework that jointly predicts NOCS correspondence, object pose and size, and a canonical Semantic Gaussian Field. Rather than treating reconstruction as a detached auxiliary task, \ours{} uses the Gaussian field as a category-conditioned structural prior that participates in multimodal feature fusion and provides global guidance for local pose reasoning. Conditioned on a categorical text embedding, \ours{} processes RGB-D observations through a transformer-based fusion module that integrates visual, geometric, and category-level cues, decoding the NOCS map, pose and size information and the Gaussian-based object representation. Extensive experiments on synthetic and real-world benchmarks show that this coupled pose-shape formulation achieves strong performance in a single-model multi-category setting, with up to 22\% on SOPE and competitive transfer to real-world data. These results highlight the benefit of jointly learning canonical correspondence, object shape, and pose alignment within a unified representation.
\end{abstract}

\begin{figure}[H]
  \centering
  \includegraphics[width=\linewidth]{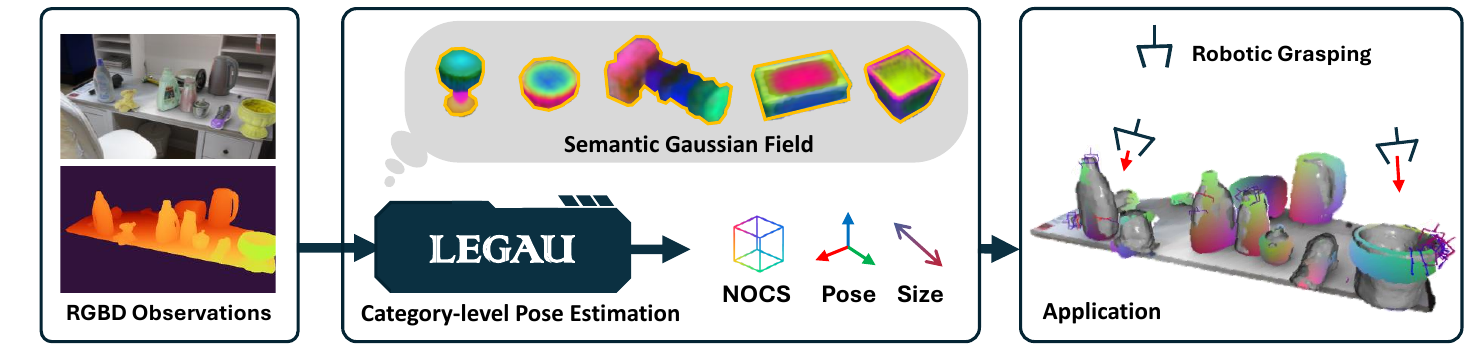}
  \vspace{-0.15in}
  \caption{\small
    We present \textbf{\ours}, a unified framework for scalable category-level
    object pose estimation from a single view. Our method learns a
    \textit{Semantic Gaussian Field} that provides shape- and category-aware
    guidance for coordinate alignment, enabling simultaneous object understanding
    and pose estimation, which may support downstream robotic perception tasks.
  }
  \label{fig:teaser}
  \vspace{-0.1in}
\end{figure}

%-------------------------------------------------------------------------
\section{Introduction}
% A core capability of modern embodied AI is to interpret its surroundings with human-like priors and common-sense reasoning. Robust and generalized object pose estimation is particularly essential for reliable robotic manipulation. Recent methods for pose estimation on unseen objects~\cite{foundationpose,ornek2024foundpose,matchu,sam6d,moon2024genflow,osop} have shown strong progress, but they rely on accurate CAD models, limiting their applicability in real-world settings. Category-level pose estimation~\cite{wang2019normalized,chen2021fsnet,di2022gpv,li2025gce,zheng2023hs} provides a model-free alternative by learning a canonicalized representation shared across instances within a category. However, these approaches still struggle to generalize across substantial intra-class variation in shape and appearance. Incorporating category-level priors to achieve robust, instance-agnostic pose estimation with human-like perceptual abilities remains an open challenge.
\hongli{Category-level 6D pose estimation aims to recover the pose of previously unseen object instances by leveraging shared structural priors across object categories. 
While recent methods have shown promising progress~\cite{wang2019normalized,chen2021fsnet,di2022gpv,li2025gce,zheng2023hs}, they often struggle to generalize across substantial intra-class variation in shape and appearance. 
A key limitation is that pose estimation and shape understanding are typically modeled as separate processes: pose is inferred from partial observations, while shape or semantic priors are introduced only as auxiliary signals. Without a unified representation that jointly captures object geometry and semantics, these approaches remain sensitive to occlusion, incomplete geometry, and large shape variation.}

\hongli{Several recent works attempt to incorporate stronger priors for pose estimation.} Some methods~\cite{lin2024instance,chen2024secondpose} integrate visual semantics from DINOv2~\cite{oquab2023dinov2} with geometric features from PointNet~\cite{qi2017pointnet} like encoders to jointly reason about object pose and shape from partial RGB(D) observations. 
In order to have access to global context in heavy occlusion scenarios, recent work~\cite{cai2025gs, li2025gce} further introduces global context priors or canonical prototypes to guide the pose estimation, but they suffer from per-category training for shape reconstruction, limiting their scalability and generalization to new categories.
Another line of research estimates object pose by first reconstructing object geometry and then performing model-based alignment~\cite{sun2022onepose,he2022oneposeplusplus,foundationpose}. 
More recently, scholars~\cite{nguyen2024gigaPose,lee2025any6d,geng2025one} leverage large image-to-3D models~\cite{xiang2025structured,ye2025hi3dgen} to synthesize object shapes before performing model-based pose estimation~\cite{foundationpose,nguyen2024gigaPose,ornek2024foundpose}. 
While these methods produce visually accurate reconstructions, their performance heavily depends on visible regions; under partial or occluded observations, the incomplete geometry often leads to pose misalignment and degraded accuracy.

% Earlier work introduces global shape priors by decoupling object reconstruction from pose estimation. However, from a single RGB(D) observation, pose estimation implicitly aligns an object model with the input, meaning that object reconstruction as well as its semantic understanding are inherently entangled with pose reasoning. This raises a natural question: can we unify these coupled components within one framework, enabling shape and semantic priors to directly guide pose alignment?
\hongli{These limitations stem from a fundamental issue: pose estimation implicitly requires aligning an object representation with the observed input, meaning that object reconstruction, semantic understanding, and pose reasoning are inherently coupled. 
This raises a natural question: \textit{can shape, semantics, and pose reasoning be unified within a single representation that directly supports pose alignment?}
}

We argue that category-level pose estimation should be viewed as a coupled pose-shape inference problem rather than an isolated pose regression task. 
Motivated by this, we present \ours{}, a unified framework for 6D pose estimation guided by a Semantic Gaussian Field that jointly supports NOCS prediction, pose estimation, and Gaussian-based shape reconstruction. \ours{} learns a latent field embedding as a category-conditioned structural prior. This prior couples pose reasoning and shape reconstruction by allowing local RGB-D evidence to interact with a canonical object-level representation. Built on a multimodal Transformer architecture, \ours{} fuses visual-geometric local embeddings, text embeddings, and latent field embeddings to jointly learn canonical correspondence, object shape, and pose. Experiments on synthetic and real-world category-level datasets show strong performance in a single-model multi-category setting, while also revealing remaining challenges in synthetic-to-real transfer.

% \begin{itemize}
%     \item We present \textbf{\ours{}} as a unified and scalable framework for category-level 6D pose estimation that presents semantic reconstruction-assisted NOCS and pose prediction within an end-to-end architecture.

%     \item We design a multi-modal pipeline that fuses local visual–geometric cues, textual category priors, and global field embeddings, achieving large-scale generalization for simultaneous object shape understanding and coordinate alignment.  

%     \item We propose a Semantic Gaussian Field that integrates semantic cues with the reconstructed Gaussian, providing contextual priors for robust pose estimation and enabling object-level understanding for robotic manipulation.  

% \end{itemize}
\hongli{
% \begin{itemize}
% \item We introduce \ours{}{}, a unified framework that integrates category-level pose estimation and object reconstruction through a shared semantic field representation.

% \item We propose a \textbf{Semantic Gaussian Field} that jointly models object geometry and semantics, providing a structured prior for robust pose alignment across unseen instances.

% \item We develop a multimodal pipeline that combines visual, geometric, and textual priors within a transformer architecture to enable scalable category-level pose estimation.
% \end{itemize}}
\begin{itemize}
\item We formulate category-level pose estimation as a coupled inference problem over canonical correspondence, object-level shape, and SE(3) pose, rather than treating pose regression and reconstruction as independent objectives.

\item We propose a Semantic Gaussian Field that models canonical object geometry together with category-conditioned feature-level cues, providing a structured prior for pose-shape coupling under partial observations.

\item We develop a multimodal transformer pipeline in which local RGB-D features, category-level text cues, and field embeddings interact through attention, enabling the canonical field to provide global structural guidance for local pose reasoning.
\end{itemize}}
% \vspace{-0.2cm}

% \begin{figure}[t] % 't' option to place at the top
%     \hspace*{\fill} % Push the figure to the right
%     \begin{minipage}[t]{0.46\textwidth} % Half-width of the text
%         \centering
%         \includegraphics[width=0.9\textwidth]{pic_\ours{}/\ours{}_teaser.png} % Adjust file name
%         \caption{\ours{} canonically aligns Objaverse shapes to OmniNOCS and unifies NOCS prediction, pose estimation, and neural reconstruction from both real and synthetic inputs.}
%         \vspace{-0.5cm}
%         \label{fig:teaser}
%     \end{minipage}
% \end{figure}

\section{Related Work}

\paragraph{Category-Level Object Pose Estimation.}
Estimating the 9DoF pose of objects without relying on instance-specific CAD models has been a core challenge in object-centric perception. 
Early works focused on \textit{instance-level} settings~\cite{drostppf,wang2021gdr,foundationpose,sam6d,matchu,megapose}, where pose could be recovered by registering observed point clouds to known meshes using geometric optimization such as ICP or PnP. 
However, such methods do not generalize to novel instances or categories. 
To overcome this, \textit{category-level pose estimation} introduces the concept of a Normalized Object Coordinate Space (NOCS)~\cite{wang2019normalized}, where each object instance is aligned to a canonical coordinate frame shared across a category. 
Earlier methods ~\cite{chen2020learning,chen2021fsnet,lin2021dualposenet,di2022gpv,diffusionnocs} follow this pose representation  and opt to introduce powerful backbones to improve the regression of the NOCS maps, as this formulation enables generalization to unseen instances.
Recent approaches extend this framework with semantic and geometric priors. 
SGPA~\cite{chen2021sgpa} introduces shape-guided priors for improving geometric stability under occlusion. 
HS-Pose~\cite{zheng2023hs} and IST-Net~\cite{liu2023net} refine the NOCS predictions using hierarchical attention or implicit surface representations, achieving better local alignment but still treating shape and pose as two independent branches. 
GenPose~\cite{zhang2023genpose}, GenPose++~\cite{zhang2024omni6dpose} and GCE-Pose~\cite{li2025gce} further incorporate shape supervision from synthetic data or voxelized reconstructions, yet these designs rely on explicit canonical models and do not fully couple the reconstruction with pose reasoning. 
In contrast, our \ours{} framework unifies NOCS regression and object shape understanding in a shared SE(3)-aware feature space, where both tasks co-evolve through mutual attention, which leads to semantic and geometric consistency.

\noindent \textbf{Object Reconstruction for Pose Estimation.}
A natural way to handle object pose estimation when the object model is unavailable is to first reconstruct the object.
Previous research has explored several strategies for modeling objects using different representations.
OnePose~\cite{sun2022onepose}, OnePose++\cite{he2022oneposeplusplus}, and CosyPose\cite{labbe2020cosypose} reconstruct objects via Structure-from-Motion (SfM)\cite{sfm} from multi-view inputs.
More recently, neural radiance fields (NeRFs)\cite{nerf} have enabled continuous shape representations with high-fidelity view synthesis.
NeRF-based approaches~\cite{yen2020inerf,li2023nerf,foundationpose} model view-dependent appearance and achieve realistic image synthesis, effectively reducing domain gaps for pose estimation in real-world environments.
Meanwhile, 3D Gaussian Splatting (3D-GS)~\cite{kerbl20233d} provides an efficient, differentiable, and 3D-consistent representation that benefits downstream reasoning tasks such as pose estimation.
Recent works such as GS-Pose~\cite{cai2025gs}, 6D-GS~\cite{matteo20246dgs}, and 6DOPE-GS~\cite{Jin_2025_6dopgs} leverage 3D-GS for pose estimation, yet they lack category-level priors and can only reconstruct visible regions rather than complete object shapes. Benefit from the 3D generative models, GigaPose~\cite{nguyen2024gigaPose} and Any6D~\cite{lee2025any6d} opt to generate the object 3D within a reference image, while the generated model is used to support the model-based pose estimation. These approaches separate pose estimation from reconstruction and depend on generated shapes that may hallucinate geometry or appearance. In contrast, our method directly predicts Gaussian primitives. Instead, our approach builds an end-to-end pipeline that uses a compact semantic Gaussian representation, removing per-instance optimization, and guides pose alignment from partial observations.

\noindent \textbf{Joint Learning of Object Shape and Pose.}
Unifying pose estimation and shape reconstruction has recently gained attention as researchers seek SE(3)-consistent object understanding. 
GCE-Pose~\cite{li2025gce} introduces geometry-conditioned embeddings for better alignment between NOCS and 3D shapes. 6D-GS~\cite{matteo20246dgs}
is capable of modeling the 3D-GS as well as 6D pose from a single frame, and 6DOPE-GS~\cite{Jin_2025_6dopgs} introduces an online pipeline to perform live object pose tracking and reconstruction, while they lack global priors for a complete shape generation, limited by the viewpoints in observation.
Nevertheless, most of these frameworks rely on two-stage or iterative optimization pipelines, where pose and shape are refined alternately rather than jointly. 
A closely related recent direction is category-agnostic pose-shape estimation. ~\cite{zhang2025beyondtemplates} jointly predicts pose, size, and dense shape from a single RGB-D image without test-time templates, CAD models, or category labels, using foundation-model features, partial point clouds, and a MoE-enhanced Transformer. In contrast, LEGAU does not primarily pursue category-agnostic inference; instead, it studies category-conditioned pose-shape coupling, where a Semantic Gaussian Field acts as a canonical structural prior that interacts with local RGB-D features for NOCS prediction and pose alignment.

In summary, prior works either tackle category-level pose estimation but lack holistic 3D reasoning, resulting in limited global guidance across object instances, or focus on high-fidelity reconstruction while overlooking spatial and semantic alignment between the reconstructed shape and the partial observation. \ours{} unifies these two directions through a multi-modal transformer that jointly leverages visual, geometric, and semantic cues. By coupling NOCS prediction, 6D pose regression, and semantic Gaussian field reconstruction within a single framework, our method encourages SE(3)-consistent object representations within a single shared model, avoiding per-category reconstruction networks.

% \todo{diffusion based joint pose and shape(orientation matters)}
% % \paragraph{Discussion.}
% % In summary, prior works either (1) focus on category-level pose estimation but lack holistic 3D reasoning, or (2) emphasize high-fidelity reconstruction but ignore spatial alignment. 
% % \ours{} bridges these two research directions through a unified multi-modal transformer architecture that reasons jointly over visual, geometric, and semantic cues. 
% % By coupling NOCS prediction, 6D pose regression, and Gaussian-based reconstruction with differentiable rendering refinement, our method achieves SE(3)-consistent object understanding that generalizes across both synthetic and real-world domains.

\section{Method}
\subsection{Overview}
\ours{} takes a partially observed RGB-D object image as input, conditioned on a categorical text prompt, and predicts both the NOCS map and a complete semantic Gaussian field within a unified transformer architecture. The framework integrates three complementary embeddings: a text embedding extracted using CLIP~\cite{radford2021learning}, a local embedding derived from RGB features (DINOv2~\cite{oquab2023dinov2}) and depth points (PointNet~\cite{qi2017pointnet}), and a field embedding that encodes global semantic shape priors. \hongli{The multimodal embeddings are fused through alternating self-attention and cross-attention layers. Self-attention aggregates information within each modality, while cross-attention enables interaction between local observations, textual priors, and the global semantic field. This design allows the model to propagate global structural cues to local pose reasoning while preserving fine-grained geometric information.}

\begin{figure*}[t]
    \centering
    \includegraphics[width=\textwidth]{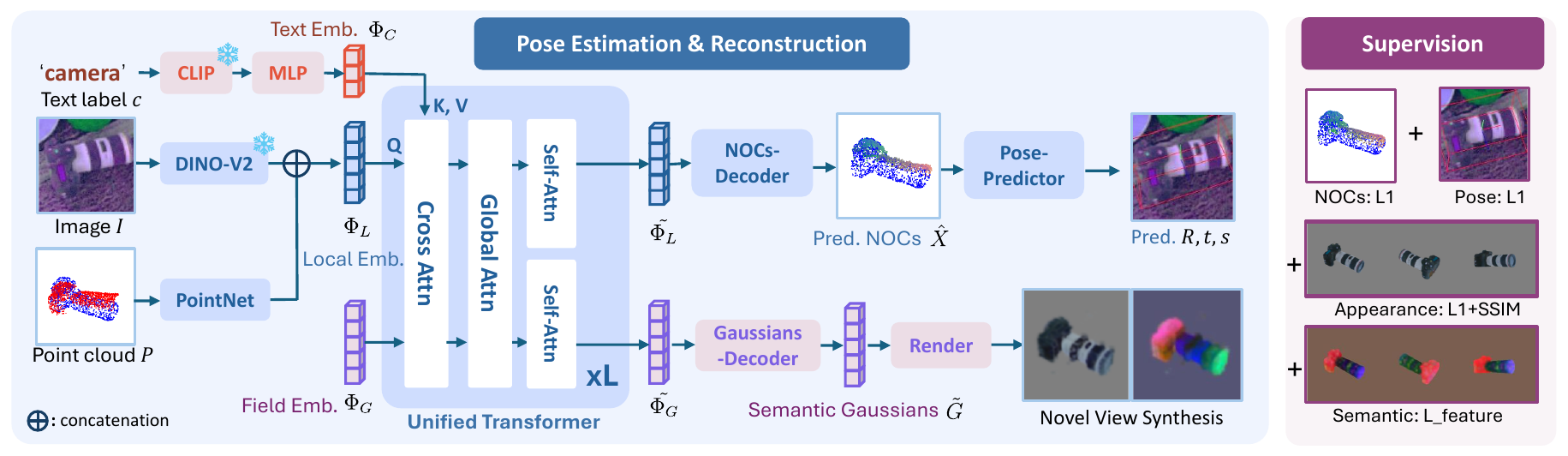}
    \vspace{-0.7cm}
    \caption{
         % \todo{(1)specify dino vx; (2)enlarge the object image, (3)adjust the input depth points(we take depth points cloud as input instead of depth image. make it consistent; (4)rename the embeddings, (5)mark on the backbones if they are frozen.(6) add a "+" mark between dino and pointnet and specify if its concatenate or adding operation.}
        \textbf{Pipeline of the \ours{} framework.} Given an RGB-D input and a categorical text prompt, \ours{} extracts modality-specific features using DINOv2~\cite{oquab2023dinov2}, CLIP~\cite{radford2021learning}, and PointNet~\cite{qi2017pointnet++}, producing a \textit{text embedding} that encodes category priors and a \textit{local embedding combining} visual and geometric cues. A learnable \textit{field embedding} captures global semantic and geometric priors and jointly attends with the multi-modal embeddings through a transformer.
        This unified representation enables coupled semantic shape understanding and pose reasoning: local embeddings drive the NOCS and pose decoders, while the global embedding guides the shape decoder to reconstruct high-level semantics and geometry as Gaussian primitives. The training of our model uses multi-level supervision, including NOCS and pose losses, photometric consistency, and a cosine-similarity objective for feature coherence.}      
    % \vspace{-0.7cm}
    \label{method:pipeline}

\end{figure*}

\subsection{Task Formulation.}
\noindent Given a cropped RGB image $I \in \mathbb{R}^{H \times W \times 3}$ and its corresponding partial point cloud $P \in \mathbb{R}^{N \times 3}$ obtained from depth observations, the goal of \ours{} is to jointly estimate the object’s 6D pose $\{R, t\} \in \mathrm{SE}(3)$, its 3D size $s \in \mathbb{R}^3$, the normalized object coordinates $\hat{X} \in \mathbb{R}^{N \times 3}$, and a dense feature Gaussian representation $\mathcal{G}$ in the canonical space.

\subsection{Preliminary: Feature 3D Gaussians}
\label{sec:3dfgs}

We follow the 3D Feature Gaussians formulation~\cite{zhou2024feature} and represent an object using a set of Feature Gaussian primitives
$\mathcal{G}=\{G_k\}_{k=1}^{K}$. 
Each primitive $G_k$ stores geometric, appearance and semantics parameters,
\[
G_k = (\mu_k,\;\Sigma_k,\;\alpha_k,\;c_k,\;f_k),
\]
where $\mu_k\!\in\!\mathbb{R}^3$ is the Gaussian center, 
$\Sigma_k\!\in\!\mathbb{R}^{3\times 3}$ the anisotropic covariance, 
$\alpha_k$ the opacity, 
$c_k$ the color feature, 
and $f_k\!\in\!\mathbb{R}^{C_f}$ a learnable feature embedding. 

\noindent \textbf{Semantic Gaussian Field.}
% Rather than following the traditional ``first reconstruct, then regress pose’’ paradigm, our goal is to learn pose and shape \emph{simultaneously} within a shared representation.    
% We therefore introduce a learnable \textit{Semantic Gaussian Field}—a triplane-based latent embedding that captures canonical geometry, semantics, and part structure before the explicit Gaussians are decoded.  
% This field serves as an implicit prior that informs pose estimation throughout the network, enabling the model to understand object structure, resolve ambiguities, and remain consistent under occlusion or partial observations.  
\hongli{
Rather than following the traditional ``reconstruct-then-align'' paradigm, LEGAU learns object pose and shape within a shared representation. We introduce a learnable \textbf{Semantic Gaussian Field}, a triplane-based
latent embedding that captures canonical geometry, semantics, and part structure before explicit Gaussian primitives are decoded.
Importantly, the semantic field directly influences pose reasoning. During transformer attention, the field embedding interacts with local visual and geometric features, providing global structural cues that guide NOCS
prediction and pose estimation.
This allows the model to resolve ambiguities caused by occlusion or partial observations by enforcing consistency with the learned canonical object structure.
Noticeably, “semantic” refers to category-conditioned and feature-level structural cues encoded by text and visual foundation-model features, rather than explicit part labels or functional language semantics.
}

\subsection{Multimodal Representation}
\label{sec:encoding}
\ours{}{} learns a unified multimodal representation that integrates complementary cues from visual, geometric, and text sources. 
Given an RGB image $I \in \mathbb{R}^{H \times W \times 3}$, a depth-derived point cloud $P \in \mathbb{R}^{N \times 3}$, and a category label $c$, 
the model extracts modality-specific embeddings and fuses them into a shared latent space.
% \xu{consider the name for different kind of embeddings}

\noindent\textbf{{Local Embedding:}}  A pre-trained foundational encoder~\cite{oquab2023dinov2} backbone extracts dense visual features 
$F_{\text{rgb}} = \mathcal{F}_{\text{vis}}(I) \in \mathbb{R}^{H' \times W' \times C}$ 
that provide semantically consistent appearance cues across categories and viewpoints. 
To encode geometric structure, a lightweight point encoder~\cite{qi2017pointnet} processes the partial point cloud 
$P$ to obtain geometric embeddings 
$F_{\text{pc}} = \mathcal{F}_{\text{geo}}(P) \in \mathbb{R}^{N \times C}$ 
representing SE(3)-aware local 3D geometry.
We then sample and align corresponding spatial locations from $F_{\text{rgb}}$ and $F_{\text{pc}}$, producing a unified set of tokens $\phi_L \in  \mathbb{R}^{K \times C_f}$, each embedding both appearance and geometry in a shared latent space. These tokens form the local embedding, which provides spatially grounded cues for NOCS prediction 
and simultaneously incorporates with the global field embedding, enabling the transformer to aggregate coherent shape and semantic context across views and modalities.

\noindent\textbf{Text Embedding:}  
To incorporate language-level priors, we embed the object’s category label $c$ using a text encoder~\cite{radford2021learning}, 
yielding a semantic token $\phi_C \in \mathbb{R}^{1 \times C}$. 
This categorical embedding provides high-level contextual guidance that complements the local visual–geometric cues, 
allowing the model to infer missing structure under occlusion or sparse observations. 

\noindent\textbf{Field Embedding:}  
Beyond input encoding, \ours{} introduces a learnable field embedding 
$\phi_G \in \mathbb{R}^{3 \times H_f \times W_f \times C_f}$, 
that captures holistic geometry and semantic context. It is parameterized as a compact triplane feature
$\Phi_G = \{\Phi_{xy}, \Phi_{yz}, \Phi_{zx}\} \in \mathbb{R}^{3 \times H_f \times W_f \times C_f}$, 
where each orthogonal plane encodes a 2D projection of the latent 3D structure. 
This field embedding serves as a geometry-aware semantic prior, 
providing global structural guidance and regularizing pose reasoning for consistent alignment across unseen categories.

\subsection{Unified Transformer Backbone}
\label{sec:backbone}
\noindent
To unify visual, geometric, and semantic information across modalities, 
\textit{LEGAU} employs a transformer-based backbone that progressively fuses and contextualizes the input tokens defined in Section~\ref{sec:encoding}. 
Inspired by recent large-scale geometric transformers such as VGGT~\cite{wang2025vggt}, we employs an alternating-attention backbone that unifies reasoning across multiple embedding groups. This design enables effective fusion of visual, geometric, and semantic cues, while maintaining stable optimization and strong generalization across object categories.

\noindent{\textbf{Token Grouping.}}
The input token set is divided into four groups:  
(i) three triplane field embeddings 
$\Phi_{xy}, \Phi_{yz}, \Phi_{zx} \in \mathbb{R}^{K_F \times C_F}$,  
each encoding 2D projections of the latent 3D feature field along orthogonal planes;  
(ii) a local embedding 
$\phi_L \in \mathbb{R}^{K_L \times C}$ capturing spatially grounded visual–geometric features.  
Together, these tokens are concatenated into a unified multimodal sequence 
$T = \{\Phi_{xy}, \Phi_{yz}, \Phi_{zx}, \phi_L\}$ 
and processed jointly within the transformer.

\noindent{\textbf{Alternating Attention.}}
Each transformer block interleaves three attention operations—global, local, and category-conditioned—applied sequentially to $T$:
\begin{equation}
\begin{aligned}
T' &= \text{GlobalAttn}\big(\text{LN}(T)\big) + T, \\
T'' &= \text{LocalAttn}\big(\text{LN}(T')\big) + T', \\
T^{+} &= \text{CrossAttn}\big(\text{LN}(T''), \phi_C\big) + T'',
\end{aligned}
\end{equation}
where global attention exchanges information across triplane and local groups, 
local attention refines spatial coherence within each feature set, 
and cross-attention injects semantic priors from $\phi_C$ into all representations.
Stacking $L$ alternating-attention layers enables rich bidirectional interaction between 
local and global contexts. 
The triplane embeddings gradually consolidate multi-view geometric structure and high-level semantics, 
while local tokens preserve SE(3)-aware spatial detail for fine alignment. 
Through repeated category-conditioned attention, semantic priors are continuously propagated across modalities,  yielding geometry-aware and semantically consistent feature representations 
for downstream NOCS and pose estimation.

\noindent{\textbf{Output Representation.}}
After $L$ layers, the transformer outputs refined local embeddings $\tilde{\phi}_L$ 
and a global field embedding $\tilde{\Phi}$, 
which together capture geometry-aware priors and spatially aligned features 
for Gaussian field reconstruction and 6D pose estimation.

\subsection{Gaussians Decoder}
\label{sec:shape_decoder}

\noindent
The Gaussians Decoder reconstructs the object’s geometry by sampling a set of Gaussian centers on a fixed 3D grid and decoding their attributes from the triplane field embedding $\tilde{\Phi}$. 
Unlike prior formulations that directly regress all Gaussian parameters, 
we derive per-primitive features through differentiable sampling from the triplane representation.

\noindent{\textbf{Feature Sampling.}
We uniformly sample $K$ grid centers $\{\,x_k \in \mathbb{R}^3\,\}_{k=1}^{K}$ within the canonical object volume. 
For each center $x_k$, we project it onto the three orthogonal planes of the triplane field 
$\tilde{\Phi} = \{\tilde{\Phi_{xy}}, \tilde{\Phi_{yz}}, \tilde{\Phi_{zx}\}}$, 
and aggregate its interpolated latent feature as
\begin{equation}
\psi_k = \sum_{p \in \{xy, yz, zx\}} 
      \text{Interp}(\tilde{\Phi_{p}}, x_k),
\end{equation}
where $\text{Interp}(\tilde{\Phi_p}, x_k)$ denotes bilinear interpolation on plane $\tilde{\Phi_p}$ at the projected location of $x_k$.

\noindent{\textbf{Gaussian Attribute Decoding.}}
Each latent feature $\psi_k$ is passed through a lightweight MLP decoder 
to predict the corresponding Gaussian parameters:
\begin{equation}
\{\mu_k, \sigma_k, \alpha_k, q_k, c_k, f_k\} = \text{MLP}_{\text{dec}}(\psi_k),
\end{equation}
where $\mu_k$ denotes the Gaussian center, $\sigma_k$ its anisotropic scale, 
$\alpha_k$ the opacity, $q_k$ the orientation quaternion, $c_k$ the color, 
and $f_k$ the learned feature embedding. 
This formulation allows the triplane field to compactly encode the object’s latent 3D geometry, 
while the decoder translates it into physically meaningful Gaussian attributes.

\noindent{\textbf{Differentiable Rendering.}}
Given a camera pose $T_i \in \mathrm{SE}(3)$ and the predicted set of feature Gaussians
$\tilde{\mathcal{G}}=\{\mu_k,\sigma_k,\alpha_k,q_k,f_k\}_{k=1}^{K}$,
we render to the image plane via differentiable Gaussian splatting, obtaining
view-dependent RGB, depth, and feature maps:
\begin{equation}
(\,\tilde{I_i},\,\tilde{D_i},\,\tilde{F_i}\,) \;=\; \mathrm{Render}\big(\tilde{\mathcal{G}},\, T_i\big).
\end{equation}
Here, $\tilde{F}$ is computed by accumulating the per-Gaussian feature embeddings $f_k$
along the ray compositing process, consistent with the RGB/opacity blending rules.

\subsection{NOCS and Pose Decoders}
\label{sec:nocs_pose}

\noindent
The local embeddings $\tilde{t}_L$ encode SE(3)-aware geometry and per-point semantics.  
\ours{} employs two lightweight decoders that share these features but are supervised with complementary objectives.

\noindent{\textbf{NOCS Prediction.}}
We first regress the normalized object coordinates for each input point:
\begin{equation}
\hat{X} = \mathrm{MLP}_{\text{nocs}}(\tilde{t}_L), \qquad
\hat{X} \in \mathbb{R}^{N \times 3}.
\end{equation}
This canonicalization step encourages $\tilde{t}_L$ to encode consistent geometric structure across instances.

\noindent{\textbf{Pose Feature Construction.}}
To obtain a pose-aware representation, we fuse three sources of information: local embeddings $\tilde{t}_L$, the predicted NOCS coordinates $\hat{X}$, and the depth points $P$. Each is encoded by an MLP and concatenated:
\begin{equation}
f_{\text{pose}} =
\mathrm{concat}\big[
\mathrm{MLP}(\tilde{t}_L),\,
\mathrm{MLP}(\hat{X}),\,
\mathrm{MLP}(P)
\big].
\end{equation}
% This fused feature provides strong geometric cues and reduces pose–shape ambiguity.

\noindent{\textbf{Pose and Size Estimation.}}
Rotation, translation, and anisotropic size are then regressed from $f_{\text{pose}} \rightarrow \{ R \in \mathrm{SO}(3),  t \in \mathbb{R}^3, s \in \mathbb{R}^3 \}$.
Especially, rotation is predicted using the continuous 6D representation~\cite{zhou2019continuity}, while translation and size come from independent heads:
\begin{equation}
R = \big[\hat{r}_1,\, \hat{r}_2,\, \hat{r}_1 \!\times\! \hat{r}_2\big],
\qquad
(r_1,r_2) = \mathrm{MLP}_{R}(f_{\text{pose}}),
\end{equation}
\begin{equation}
t = \mathrm{MLP}_{t}(f_{\text{pose}}), \qquad
s = \mathrm{MLP}_{s}(f_{\text{pose}}).
\end{equation}

\subsection{Training Objectives}
\label{sec:loss}
\ours{} is trained end-to-end with a multi-task objective that jointly supervises
canonical correspondence, SE(3) alignment, and Gaussian-field reconstruction.

\noindent\textbf{NOCS and Pose Loss.}
We supervise both NOCS prediction and pose regression using a Smooth-$L_1$ formulation. 
The NOCS decoder predicts continuous normalized coordinates, and is trained with a Smooth-$L_1$ loss against the ground-truth canonical coordinates:
\begin{equation}
\mathcal{L}_{\text{nocs}}
=
\mathrm{Smooth}\text{-}L_1(\hat{X}, X^\ast).
\end{equation}
Rotation (6D representation), translation, and anisotropic size are supervised in the same manner:
\begin{equation}
\mathcal{L}_{\text{pose}}
=
L_1(R, R^\ast)
+
L_1(t, t^\ast)
+
L_1(s, s^\ast).
\end{equation}
This unified loss encourages the local embeddings to encode stable SE(3) alignment cues and consistent canonical geometry, forming a reliable foundation for downstream reconstruction and global field reasoning.

\noindent\textbf{Reconstruction Loss.}
Given the rendered RGB, depth, and feature maps
$(\tilde{I}_i, \tilde{D}_i, \tilde{F}_i)$,
we supervise the Semantic Gaussian Field using multi-view photometric and feature consistency losses:
\begin{equation}
\mathcal{L}_{\text{rgb}} =
\lVert \tilde{I}_i - I_i \rVert_1 +
\lambda_{\text{ssim}}\, \mathrm{SSIM}(\tilde{I}_i, I_i),
\end{equation}
\begin{equation}
\mathcal{L}_{\text{feat}} =
1 - \cos(\tilde{F}_i, F_i).
\end{equation}
Here, the feature supervision $F_i$ is obtained from DINOv2 features extracted on the rendered ground-truth views.

\noindent\textbf{Total Loss.}
The full training objective combines canonical correspondence, SE(3) alignment, 
and Gaussian-field reconstruction:
\begin{equation}
\begin{aligned}
\mathcal{L}_{\text{total}}
= &\ \lambda_{\text{nocs}}\, \mathcal{L}_{\text{nocs}}
+ \lambda_{\text{pose}}\, \mathcal{L}_{\text{pose}} \\
&\ + 
       \lambda_a \mathcal{L}_{\text{rgb}}
    + \lambda_f \mathcal{L}_{\text{feat}}
\end{aligned}
\end{equation}

\noindent
This multi-task formulation jointly regularizes local SE(3)-aware correspondence 
and global Gaussian-field semantics, yielding pose estimates that remain consistent 
with object geometry even under occlusion or unseen category shifts.

\section{Implementation Details}
\ours{} employs a multimodal transformer backbone with $L{=}12$ alternating-attention layers and a hidden dimension of 256. 
RGB features are extracted using a frozen DINOv2-Small encoder, while partial point clouds are processed by a lightweight PointNet with 128 hidden units. 
The Semantic Gaussian Field is represented by a triplane latent structure with three $32{\times}32{\times}32$ feature planes, and $K{=}131{,}072$ canonical Gaussian centers are uniformly sampled for reconstruction. 
We render $518{\times}518$ RGB, depth, and feature images during training using the Gaussian-splatting kernel adapted from~\cite{zhou2024feature}. 
For each object, we pre-render 42 multi-view images and randomly sample 4 views per iteration to supervise the Gaussian field during training. 
The loss weights are set to $(\lambda_{\text{nocs}}, \lambda_{\text{pose}}, \lambda_{a}, \lambda_{\text{ssim}}, \lambda_{d}, \lambda_{f}) = (2.0,\, 0.3,\, 5.0,\, 1.0,\, 0.02,\, 0.3)$ to balance the multi-objective optimization across canonical correspondence, pose regression, and Gaussian-field reconstruction. The model is trained end-to-end using AdamW with a learning rate of $2{\times}10^{-4}$, a cosine decay schedule, batch size $B{=}64$, and $200$k iterations. 
We uniformly sample $N{=}1024$ depth points per object as geometric input.

\section{Experiments}
\subsection{Datasets and Metrics}
\noindent \textbf{Datasets.}
We evaluate \ours{} across three complementary benchmarks: \textbf{HouseCat6D}~\cite{jung2024housecat6d}, \textbf{SOPE}~\cite{zhang2024omni6dpose}, and \textbf{ROPE}~\cite{zhang2024omni6dpose}, covering both real and synthetic domains. 
HouseCat6D~\cite{jung2024housecat6d} provides a compact, in-the-wild benchmark for category-level pose and shape estimation, comprising 149 real-world instances across 10 categories, including transparent objects, and featuring moderate scene clutter.
SOPE~\cite{zhang2024omni6dpose} is a large-scale synthetic dataset with 149 categories and full ground-truth annotation for pose NOCS, mesh geometry, designed to test scalability across a large number of categories. 
ROPE~\cite{zhang2024omni6dpose} shares the same 149 categories as SOPE but consists entirely of real RGB-D captures with complex lighting, background variation, and partial occlusion. 
Importantly, the model is trained only on SOPE and evaluated on ROPE without finetuning, thus it is used to measure real-world transfer performance.

\noindent\textbf{Metrics.}
For the evaluation on  Omni6DPose~\cite{zhang2024omni6dpose}, we use the metrics that proposed by this benchmark. One is \textbf{AUC@($\theta^\circ$, $\tau$cm)}, where a prediction is considered correct if the rotation error is below $\theta^\circ$ and the translation error below $\tau$ cm. We report AUC@IoU\(_{25}\), AUC@IoU\(_{50}\), and AUC@IoU\(_{75}\) for benchmarking.
The other is \textbf{VUS@\(\,n^\circ m \)cm}, which provides a fine-grained evaluation of 6D pose accuracy using the Volume Under Surface (VUS) across ranges of rotational (up to \(n^\circ\)) and translational (up to \(m\) cm) errors. It aggregates pose accuracy within these error bounds. We report VUS@5\(^\circ\)2cm, VUS@5\(^\circ\)5cm, VUS@10\(^\circ\)2cm, and VUS@10\(^\circ\)5cm. For the evaluation on the HouseCat6D~\cite{jung2024housecat6d} dataset, we follow the previous methods~\cite{lin2024instance,chen2024secondpose} to use the Accuracy on the respective \(n^\circ\) \(m\) cm as well as  the IoU metrics.

% \paragraph{HouseCat6D.}
% HouseCat6D~[?] provides category-level annotations for 6D pose and shape across 15 indoor object categories with complex geometry and cluttered backgrounds. 
% We follow the standard train/test split and report pose accuracy under angular and translational thresholds.

% \paragraph{ROPE \& SOPE.}
% ROPE~[?] and SOPE~[?] are recent large-scale synthetic benchmarks focusing on model-free pose estimation and shape understanding under occlusion and partial visibility. 
% They offer ground-truth NOCS coordinates, 6D poses, and dense mesh reconstructions across hundreds of object instances.
% We report Acc@5°2cm, Acc@5°5cm, Acc@10°2cm, and Acc@10°5cm, along with shape IoU at 25/50/75 thresholds following GenPose++~[?].

\begin{figure*}[htbp]
    \centering
    \includegraphics[width=1.0\linewidth]{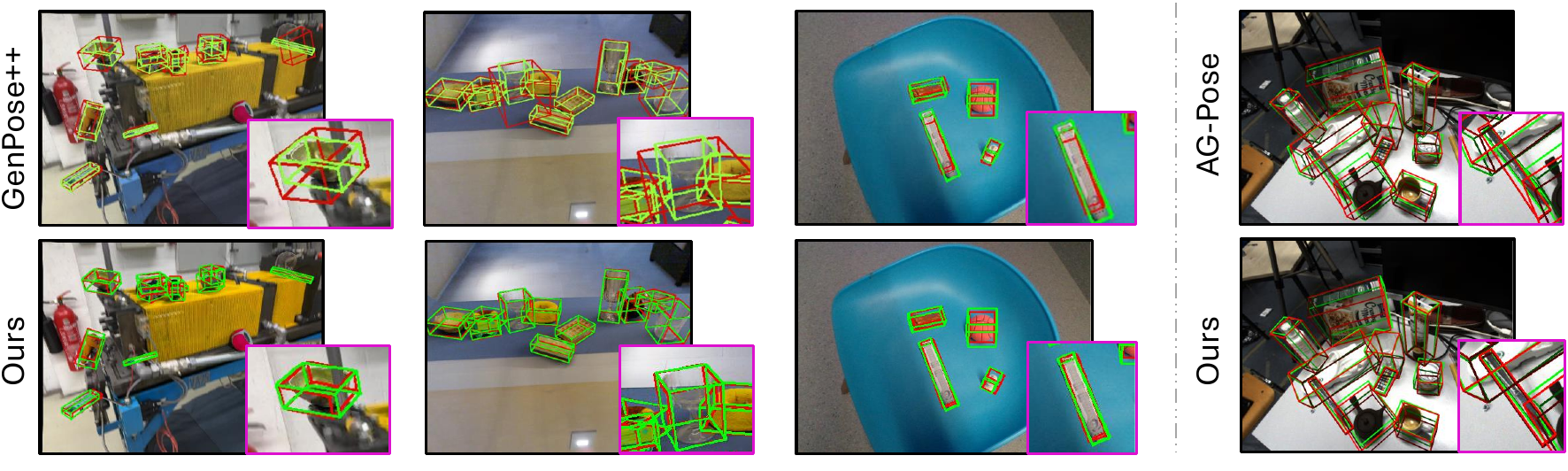}
    \vspace{-0.4cm}
    \caption{
    Qualitative comparison between our method, AGPose~\cite{lin2024instance} and GenPose++~\cite{zhang2024omni6dpose} on SOPE, ROPE, and HouseCat6D. 
    From left to right: a symmetric and occluded bowl (SOPE), a transparent glass (SOPE), a real-world toy object (ROPE), and reflective metallic tableware (HouseCat6D). 
    These examples illustrate robustness to symmetry, occlusion, transparency, real-world novel objects, and specular surfaces.
    }
    \label{fig:enter-label}\vspace{-0.4cm}
\end{figure*}

\begin{table*}[htbp]
\centering
\caption{\textbf{Quantitative comparison on HouseCat6D~\cite{jung2024housecat6d}, SOPE~\cite{zhang2024omni6dpose}, and ROPE~\cite{zhang2024omni6dpose}.} 
% We evaluate pose accuracy using IoU@25/50/75 and VUS@$n^\circ$$m$cm metrics, which jointly reflect geometric alignment under varying angular and translational tolerances. 
For each column, the top three methods are highlighted using a blue color map, where darker shades indicate better performance.
\ours{} achieves strong performance on HouseCat6D and SOPE under the same evaluation protocol, especially in the single-model multi-category setting. On ROPE, \ours{} remains competitive in IoU-based metrics but shows a gap under stricter pose thresholds, indicating that synthetic-to-real transfer remains challenging..}
\label{tab:pose_results_combined}
\resizebox{0.95\textwidth}{!}{
\begin{tabular}{l|l|ccc|cccc}
\hline
\hline
\textbf{Dataset} & \textbf{Method} &
\textbf{IoU25} & \textbf{IoU50} & \textbf{IoU75} &
\textbf{5°2cm} & \textbf{5°5cm} &
\textbf{10°2cm} & \textbf{10°5cm} \\
\hline
\hline

\multirow{4}{*}{\textbf{HouseCat6D~\cite{jung2024housecat6d}}} 
 & VI-Net~\cite{clark2017vinet}           & - & 56.4 & - & 8.4  & 10.3 & 20.5 & 29.1 \\
 & SecondPose~\cite{chen2024secondpose}  & - & \cellcolor{blue!5}66.1 & - & 
 \cellcolor{blue!5}11.0 & \cellcolor{blue!5}13.4 & 
 \cellcolor{blue!5}25.3 & \cellcolor{blue!5}35.7 \\
 & AG-Pose~\cite{lin2024instance}         & \cellcolor{blue!22}88.1 & \cellcolor{blue!22}76.9 & \cellcolor{blue!22}53.0 & \cellcolor{blue!22}21.3 & \cellcolor{blue!22}22.1 & \cellcolor{blue!22}51.3 & \cellcolor{blue!22}54.3 \\
 
 & \cellcolor{gray!35}\textbf{\ours{}}           & \cellcolor{blue!35}\textbf{90.7} & \cellcolor{blue!35}\textbf{80.6} & \cellcolor{blue!35}\textbf{57.4} & \cellcolor{blue!35}\textbf{21.4} & \cellcolor{blue!35}\textbf{22.7} & \cellcolor{blue!35}\textbf{54.0} & \cellcolor{blue!35}\textbf{57.4} \\
\hline
\hline
\textbf{Dataset} & \textbf{Method} &
\multicolumn{3}{c|}{\textbf{AUC$\uparrow$ (IoU)}} &
\multicolumn{4}{c}{\textbf{VUS$\uparrow$($n$°$m$cm)}} \\
\cline{3-9}
& &
\textbf{IoU25} & \textbf{IoU50} & \textbf{IoU75} &
\textbf{5°2cm} & \textbf{5°5cm} &
\textbf{10°2cm} & \textbf{10°5cm} \\
\hline
\multirow{7}{*}{\textbf{SOPE~\cite{zhang2024omni6dpose}}} 
 & NOCS~\cite{wang2019normalized}    & 0.0  & 0.0  & 0.0  & 0.0  & 0.0  & 0.0  & 0.0 \\
 & SGPA~\cite{chen2021sgpa}          & 13.3 & 3.2  & 0.0  & 7.7  & 10.1 & 15.0 & 20.4 \\
 & IST-Net~\cite{liu2023net}          & 36.5 & 16.9 & 1.4  & 3.6  & 5.1  & 8.6  & 11.4 \\
 & HS-Pose~\cite{zheng2023hs}           & \cellcolor{blue!5}40.1 & \cellcolor{blue!5}21.7 & \cellcolor{blue!5}3.2  & 6.3  & 8.0  & 13.6 & 17.3 \\
 & GenPose~\cite{zhang2023genpose}     & – & – & – & \cellcolor{blue!5}11.9 & 
 \cellcolor{blue!5}14.4 & 
 \cellcolor{blue!5}21.2 & 
 \cellcolor{blue!5}26.3 \\
 & GenPose++~\cite{zhang2024omni6dpose} & \cellcolor{blue!22}50.1 & \cellcolor{blue!22}31.9 & \cellcolor{blue!22}6.4  & \cellcolor{blue!22}18.4 & \cellcolor{blue!22}23.0 & \cellcolor{blue!22}31.9 & \cellcolor{blue!22}40.2 \\

 & \cellcolor{gray!35}\textbf{\ours{}}         & \cellcolor{blue!35}\textbf{61.1} & \cellcolor{blue!35}\textbf{44.7} & \cellcolor{blue!35}\textbf{15.6} & \cellcolor{blue!35}\textbf{22.6} & \cellcolor{blue!35}\textbf{26.3} & \cellcolor{blue!35}\textbf{37.4} & \cellcolor{blue!35}\textbf{44.2} \\

\hline
\multirow{7}{*}{\textbf{ROPE~\cite{zhang2024omni6dpose}}} 
 & NOCS~\cite{wang2019normalized}    & 0.0  & 0.0  & 0.0  & 0.0  & 0.0  & 0.0  & 0.0 \\
 & SGPA~\cite{chen2021sgpa}          & 10.5 & 2.0  & 0.0  & 4.3  & 6.7  & 9.3  & 15.0 \\
 & IST-Net~\cite{liu2023net}          & 28.7 & 10.6 & 0.5  & 2.0  & 3.4  & 5.3  & 8.8  \\
 & HS-Pose~\cite{zheng2023hs}           &\cellcolor{blue!5} 31.6 &\cellcolor{blue!5} 13.6 & \cellcolor{blue!5}1.1  & 3.5  & 5.3  & 8.4  & 12.7 \\
 & GenPose~\cite{zhang2023genpose}     & – & – & – & \cellcolor{blue!5}6.6  & \cellcolor{blue!5}9.6  & \cellcolor{blue!5}13.1 & \cellcolor{blue!5}\textbf{19.3} \\
 & GenPose++~\cite{zhang2024omni6dpose} & \cellcolor{blue!35}\textbf{39.0} & \cellcolor{blue!22}19.1 & \cellcolor{blue!22}2.0  & \cellcolor{blue!35}\textbf{10.0} & 
 \cellcolor{blue!35}\textbf{15.1}& 
 \cellcolor{blue!35}\textbf{19.5}& 
 \cellcolor{blue!35}\textbf{29.4}\\

 & \cellcolor{gray!35}\textbf{\ours{}}             & \cellcolor{blue!22}37.7 & \cellcolor{blue!35}\textbf{19.3} &\cellcolor{blue!35}\textbf{2.4} &\cellcolor{blue!22}7.6
 &\cellcolor{blue!22}11.7&\cellcolor{blue!22}15.6
 &\cellcolor{blue!22}24.4\\
\hline
\end{tabular}}
\end{table*}

% \subsection{Implementation Details}
% Our model is implemented in PyTorch and trained end-to-end on 8×A100 GPUs for 200 epochs.
% The input consists of a cropped RGB image of size $224\times224$, its aligned depth map converted into a partial point cloud (1,024 points), and a categorical text label encoded via CLIP~[?].
% We employ a DINOv2 encoder~[?] for visual tokens and a PointNet++~[?] backbone for geometric features.
% The transformer backbone contains 12 alternating self-, cross-, and global-attention layers with hidden dimension 512.
% The shape decoder predicts $K=256$ Gaussian primitives per instance.
% Loss terms include NOCS regression ($\mathcal{L}_{nocs}$), pose supervision ($\mathcal{L}_{pose}$), shape reconstruction ($\mathcal{L}_{shape}$), and refiner alignment ($\mathcal{L}_{ref}$):
% \begin{equation}
% \mathcal{L} = \lambda_1 \mathcal{L}_{nocs} + \lambda_2 \mathcal{L}_{pose} + \lambda_3 \mathcal{L}_{shape} + \lambda_4 \mathcal{L}_{ref},
% \end{equation}
% where $\lambda_1\!:\!\lambda_2\!:\!\lambda_3\!:\!\lambda_4 = 1\!:\!1\!:\!0.5\!:\!0.2$. 
% We use AdamW optimizer with a learning rate of $1\times10^{-4}$ and weight decay $1\times10^{-2}$.
% All modules are trained jointly without stage-wise finetuning.

\subsection{Cross-Dataset Evaluation}
\label{sec:cross_dataset}

\noindent \textbf{Results on HouseCat6D (10 categories). }
We evaluate our method on the real-world HouseCat6D~\cite{jung2024housecat6d} benchmark. \ours{} achieves the best performance across all metrics, outperforming AG-Pose~\cite{lin2024instance} by a substantial margin in both rotation and translation accuracy.
Compared to the SOTA method AG-Pose~\cite{lin2024instance}, which performs keypoint detection for the categories, \ours{} improves both rotation and translation accuracy with a large margin while providing denser and more consistent shape reconstructions. This confirms that our semantic Gaussian-guided pipeline introduces effective global priors for challenging partial inputs, even for textureless, symmetric, and occluded objects. We do not report GCE-Pose~\cite{li2025gce} in our tables, because it requires training a separate reconstruction network per category. 
This per-category training paradigm does not align with our goal of evaluating scalable, unified category-level models that share a single network across many categories.

\noindent \textbf{Scaling to SOPE (149 Categories).}
When trained on the large-scale synthetic SOPE dataset, \ours{} continues to achieve state-of-the-art performance across all metrics, with significant gains in both pose and IoU over previous models such as HS-Pose\cite{zheng2023hs} and GenPose++~\cite{zhang2024omni6dpose}. 
The model maintains high Acc@($5^\circ$,5cm) across diverse categories, including furniture, tools, and household items, demonstrating that the alternating attention backbone generalizes well to highly varied category appearances and geometries. 
The triplane-based geometry encoding further preserves fine spatial detail and improves canonical consistency across categories with large intra-class variation.

% \begin{figure}
%     \centering

%         \includegraphics[width=1.0\linewidth]{pose_single_col.png}
%     \caption{Enter Caption}
%     \label{fig:enter-label}
% \end{figure}

% \begin{figure*}[!t]
%     \centering
%     \includegraphics[width=1.0\textwidth]{pic_lunar/vis_lunar_posea.png}
%     \caption{Visualization of model-free pose estimation with posed references.}
%     \vspace{-0.3cm}
%     \label{fig:teaser}
% \end{figure*}

\noindent \textbf{Real-World Transfer: SOPE \textrightarrow\ ROPE.}
We directly evaluate the SOPE-trained model on the real ROPE dataset without any finetuning. 
On ROPE, \ours{} remains competitive in IoU-based metrics but shows a noticeable gap under stricter pose thresholds compared with GenPose++. This suggests that the proposed canonical field prior improves category-level structural consistency, while accurate real-world pose transfer remains affected by depth noise, mask quality, lighting variation, and reflective or transparent materials.

% \noindent \textbf{Discussion.}
% Across the three datasets, \ours{} consistently establishes new state-of-the-art results:
% (1) on HouseCat6D, it achieves high-precision pose and reconstruction for real small-scale categories;
% (2) on SOPE, it scales to 149 diverse synthetic categories with consistent accuracy; and
% (3) on ROPE, it transfers the learned representations to real-world scenes without retraining.
% These results highlight that \ours{}’s unified design—combining local SE(3)-aware tokens, global shape embeddings, and differentiable rendering refinement—enables effective generalization from compact real settings to large synthetic corpora and back to real-world deployment.

% \begin{figure*}[t]
%     \centering
%     \includegraphics[width=\textwidth]{image_pose.png}

% %    \vspace{-0.7cm}
%     \caption{Visualization of Canonicalized shape from Objaverse}
%  %   \vspace{-0.3cm}
%     \label{fig:pipeline_overview}
% \end{figure*}

% \begin{figure}[h]
%     \centering
%     \includegraphics[width=1.0\linewidth]{image_recon.png}
%     \caption{Enter Caption}
%     \label{fig:enter-label}
% \end{figure}

\begin{figure}[t]
    \centering
    \includegraphics[width=1.0\columnwidth]{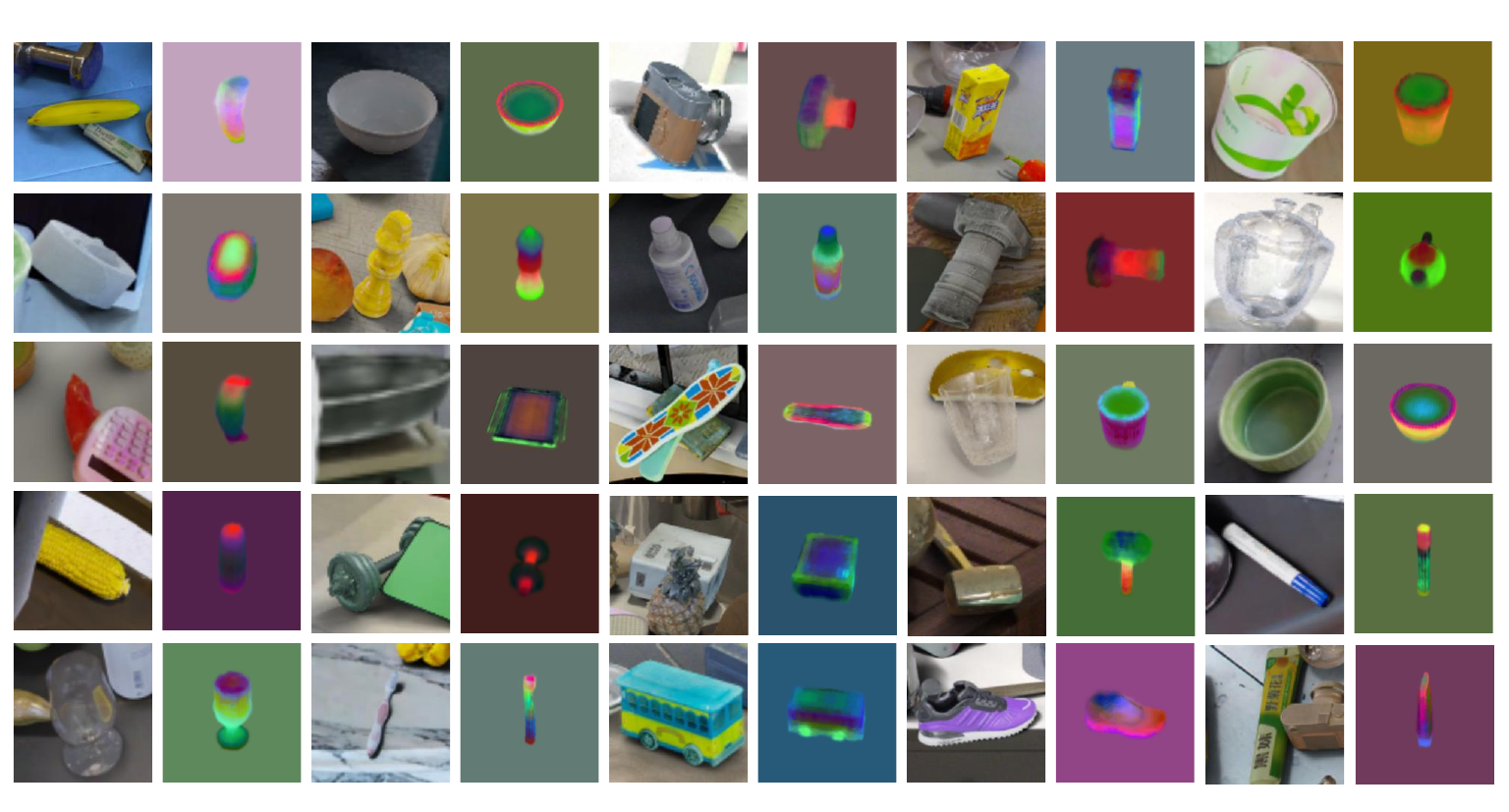}
    \caption{
        Visual of Semantic Gaussians Reconstruction.It renders Gaussian primitives decoded in canonical object space, where the per-Gaussian semantic feature embeddings are projected
to RGB colors using PCA.
    }
    % \vspace{-0.3cm}
    \label{sem_recon_results}
\end{figure}

\subsection{Effect of Learning Semantic Priors}
\noindent\textbf{Visualizing the Learned Semantic Gaussian Field.}
Before analyzing the quantitative contribution of semantics, we first provide
a qualitative visualization of the learned Semantic Gaussian Field (Fig.~\ref{sem_recon_results}). Across diverse object categories, we observe that the learned field exhibits
strong structural consistency. Regions that correspond to similar functional
or geometric structures tend to share similar colors in the latent space,
while geometrically distinct parts are clearly separated.
For elongated objects, the semantic features form smooth axial gradients;
for symmetric objects, the embeddings align around their canonical symmetry
axes. These patterns indicate that the learned representation captures
category-level structural priors rather than merely local appearance cues.

Importantly, these semantic embeddings are learned jointly with pose estimation
and shape reconstruction. As a result, the semantic field provides a global
structural prior that helps align partial observations with plausible canonical
object configurations, which stabilizes pose estimation under occlusion or
incomplete geometry.

% \begin{figure}[t]
%     \centering

%     \includegraphics[width=1.0\textwidth]{pic/contrastive_vis.pdf}
%     \caption{Gaussians reconstruction and pose estimation result under different text conditioning.}

%     \label{fig:ablation_qual}
% \end{figure}

\begin{table*}[t]
\centering

\caption{\footnotesize Ablations on isolating the effects of Gaussian reconstruction (left) and CLIP-based category conditioning (right).}

\begin{minipage}[t]{0.485\linewidth}
\centering

\resizebox{\linewidth}{!}{
\begin{tabular}{cc|cc}
\hline
\textbf{GF} & \textbf{Sem.\ GF} &
\textbf{10°5cm} & \textbf{IoU50} \\
\hline
\rowcolor{gray!15}
\cmark & \cmark & \textbf{57.4} & \textbf{80.6} \\
\cmark & \xmark & 55.1 (\textcolor{red!50!black}{-4.4\%})  & 78.3(\textcolor{red!50!black}{-2.9\%}) \\
\xmark & \xmark & 52.0(\textcolor{red!50!black}{-9.4\%}) & 75.0 (\textcolor{red!50!black}{-7.0\%})\\
\hline
\end{tabular}}
\vspace{-1mm}
% {\footnotesize \textbf{HouseCat6D}}
\end{minipage}
\hfill
\begin{minipage}[t]{0.45\linewidth}
\centering
\resizebox{\linewidth}{!}{
\begin{tabular}{c|cc}
\hline
\textbf{CLIP} &
\textbf{10°5cm} & \textbf{IoU50} \\
\hline
\rowcolor{gray!15}
\cmark & \textbf{44.2} & \textbf{44.7} \\
\xmark & 40.7(\textcolor{red!50!black}{-7.9\%}) & 42.4(\textcolor{red!50!black}{-5.1\%}) \\
\hline
\end{tabular}}
\vspace{-1mm}
% {\footnotesize \textbf{SOPE}}
\end{minipage}

\label{tab:semantic_recon_clip}
\end{table*}

\noindent\textbf{Isolated Contribution of Semantic Priors.}
To isolate the effect of semantic priors from geometry reconstruction,
we perform controlled ablations on the Semantic Gaussian Field (Sem.\ GF)
and CLIP-based category conditioning while keeping all other components unchanged.
As shown in Table~\ref{tab:semantic_recon_clip},
removing the Semantic Gaussian Field while keeping the Gaussian reconstruction branch (GF)
reduces pose accuracy on HouseCat6D from 57.4 to 55.1 (-4.4\%),
indicating that the improvement does not solely come from geometry reconstruction
but from the semantic field that encodes category-level structural priors.
Similarly, removing CLIP-based conditioning decreases VUS@10°5cm on SOPE from
44.2 to 40.7 (-7.9\%), suggesting that language-guided semantic anchors provide
useful global context under large category diversity.

% \noindent\textbf{How Semantics Improve Performance.}
% Fig.~\ref{fig:ablation_qual} shows a controlled intervention where the RGB-D input is fixed
% while only the category text prompt is changed.
% Correct prompts yield category-consistent semantic fields and accurate pose alignment,
% whereas mismatched prompts bias the reconstruction and degrade pose estimation (red boxes).
% This indicates that text-conditioned semantics acts as a global structural prior
% that reduces pose ambiguity under partial observations.

\begin{figure}[t]
    \centering
    \includegraphics[width=0.95\columnwidth]{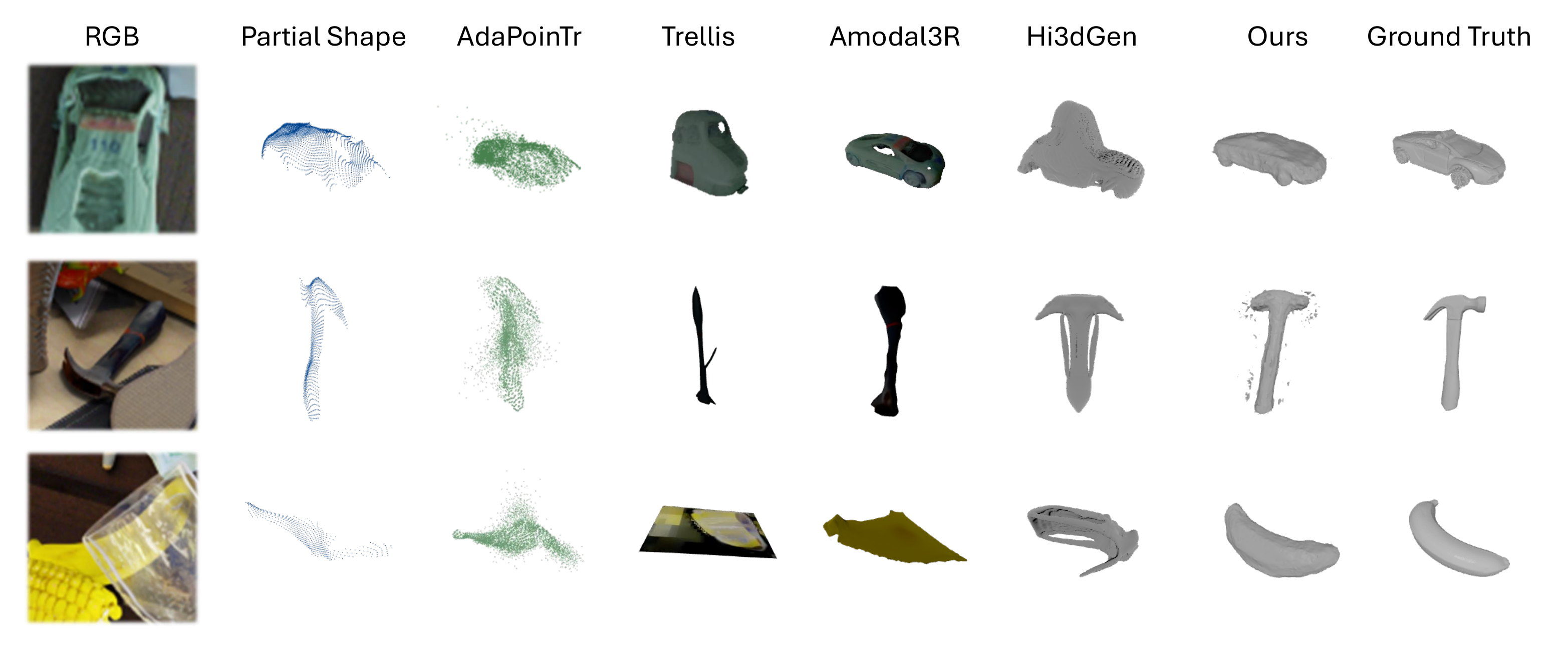}
    \caption{
    Qualitative comparison of shape reconstruction from partial point observations. 
    We compare our method with a depth-based reconstruction baseline~\cite{yu2301adapointr} and recent image-to-3D generative frameworks~\cite{xiang2025structured,wu2025amodal3r,ye2025hi3dgen} under challenging scenarios involving heavy occlusion and transparent materials.
    }
    % \vspace{-0.3cm}
    \label{recon_results}
\end{figure}

\begin{table}[h]
\centering
\caption{Shape reconstruction results on the \textbf{SOPE} dataset. 
We evaluate on the \textbf{Chamfer-L1} ($\times10^{-3}$m, \textbf{lower is better}). 
Best results are \textbf{bolded}.}
\label{tab:sope_recon}
\resizebox{0.8\linewidth}{!}{%
\begin{tabular}{cccc}
\hline
\hline
 \textbf{FoldingNet}~\cite{yang2018foldingnet} & 
\textbf{PoinTr}~\cite{yu2021pointr} & 
\textbf{AdaPoinTr}~\cite{yu2301adapointr} & 
\cellcolor{gray!15}\textbf{Ours} \\
\hline
 62.72 & 29.87 & 23.17 & \cellcolor{gray!15}\textbf{6.72}(\textcolor{green!50!black}{-16.5\%}) \\
\hline
\end{tabular}}
\end{table}

\subsection{Shape Reconstruction Results.}
Besides Semantic, we also evaluate \ours{}'s capability for dense 3D shape reconstruction on the \textbf{SOPE} dataset.
Table~\ref{tab:sope_recon} reports Chamfer-L1 distances ($\times10^{-3}$, lower is better),
where \ours{} significantly outperforms previous point-based reconstruction methods such as
FoldingNet~\cite{yang2018foldingnet}, PoinTr~\cite{yu2021pointr}, and AdaPoinTr~\cite{yu2301adapointr},
achieving the lowest reconstruction error.
Fig.~\ref{recon_results} further visualizes reconstruction results and compares \ours{} with recent
image-to-3D generative models including Trellis~\cite{xiang2025structured},
Hi3DGen~\cite{ye2025hi3dgen}, and the occlusion-aware baseline Amodal3R~\cite{wu2025amodal3r}.
While these approaches can produce visually plausible shapes from canonical viewpoints,
their geometry often deteriorates under non-frontal views, heavy occlusions,
or texture-poor observations.
In contrast, \ours{} maintains consistent geometric fidelity across viewpoints,
benefiting from explicit canonical alignment and the joint learning of pose, NOCS,
and shape within a unified transformer framework.
The Semantic Gaussian Field further provides global structural priors,
allowing the model to recover more complete and structurally consistent geometry under challenging truncation or occlusion.

\section{Conclusion }

We presented LEGAU, a unified framework that treats category-level pose estimation as coupled inference over canonical correspondence, pose alignment, and Gaussian-based shape reconstruction. By using a Semantic Gaussian Field as a category-conditioned structural prior, LEGAU encourages pose estimates to remain consistent with object-level shape structure under partial observations. Experiments on HouseCat6D, SOPE, and ROPE show that this pose-shape coupling improves performance in a single-model multi-category setting, while real-world transfer remains challenging under strict pose thresholds. Controlled ablations further indicate that the Gaussian field and category-conditioned feature cues contribute to pose accuracy beyond geometry-only reconstruction. Looking forward, we believe that such coupled pose-shape representations provide a promising basis for more open-world pose estimation, where models must reason about previously unseen categories, incomplete observations, and category-level structural priors without relying on instance-specific CAD models. Extending LEGAU toward stronger open-vocabulary conditioning, broader real-world adaptation, and more interactive robotic perception remains an important direction for future work.

% \newpage
% \input{suppl}
\bibliography{egbib}
\end{document}